\documentclass[10pt,twocolumn,letterpaper]{article}

\usepackage{acpr} 
\usepackage{times}
\usepackage{caption}
\usepackage{epsfig}
\usepackage{graphicx}
\usepackage{amsmath}
\usepackage{amssymb}

\usepackage[pagebackref=true,breaklinks=true,letterpaper=true,colorlinks,bookmarks=false]{hyperref}

\acprfinalcopy 
\def\acprPaperID{120}
\ifacprfinal\fi
\begin{document}

\title{Learning Light Field Reconstruction from a Single Coded Image}

\author{Anil Kumar Vadathya$^1$, Saikiran Cholleti$^1$, Gautham Ramajayam$^2$, Vijayalakshmi Kanchana$^3$,\\ Kaushik Mitra$^1$ \\
\and
$^1$Computational Imaging Lab\\
IIT Madras, Chennai, India\\
{\tt\footnotesize \{ee15s055,ee13b021,kmitra\}@ee.iitm.ac.in}
\and
$^2$NIT Trichy\\
Tiruchinapally, India\\
{\tt\footnotesize gautham.1131@gmail.com}
\and
$^3$SSN College of Engg.\\
Chennai, India\\
{\tt\footnotesize ajithabhavana@gmail.com}
}

\makeatletter
\let\@oldmaketitle\@maketitle
\renewcommand{\@maketitle}{\@oldmaketitle
\begin{center}
\vspace{-0.8cm}
    \includegraphics[trim=0.1cm 0 0.1cm 0,clip, width=0.97\textwidth]{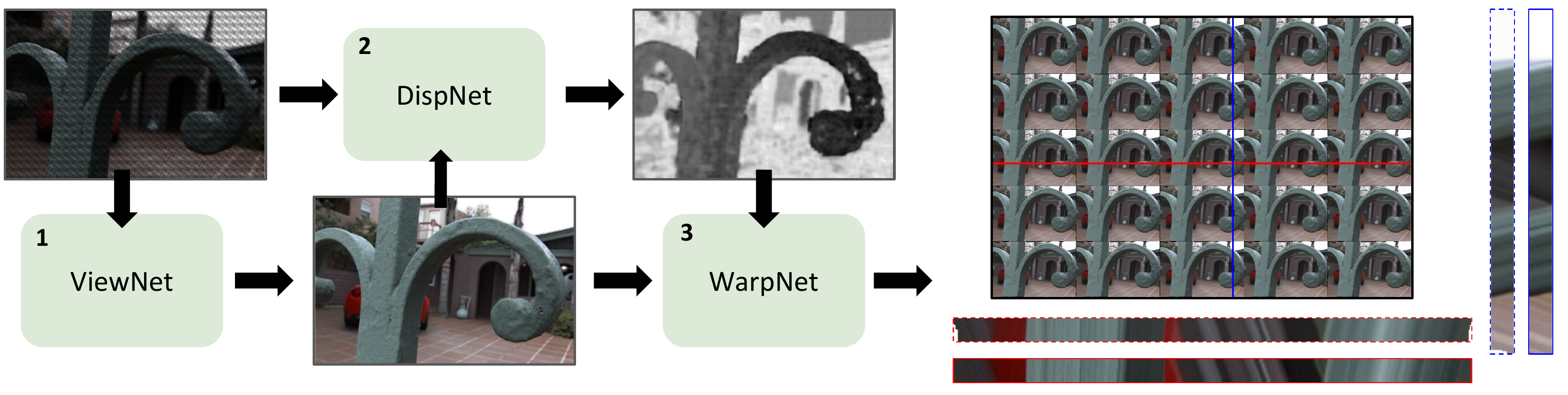}
    \captionof{figure}{We propose a learning based solution for full sensor resolution light field reconstruction from a single coded image. Our approach consists of three convolutional neural networks (CNNs); \textit{ViewNet} reconstructs center view from the coded image, \textit{DispNet} estimates disparity map, \textit{WarpNet} generates light field by warping center view with the disparity. As can be seen from EPI images, our approach recovers parallax very well. Generated EPI in dashed line border and GT in thick line.}
    \label{pipeline}
\end{center}
\bigskip}
\makeatother

\maketitle

\begin{abstract}
   Light field imaging is a rich way of representing the 3D world around us. However, due to  limited sensor resolution capturing light field data inherently poses spatio-angular resolution trade-off. In this paper, we propose a deep learning based solution to tackle the resolution trade-off. Specifically, we reconstruct full sensor resolution light field from a single coded image. We propose to do this in three stages 1) reconstruction of center view from the coded image 2) estimating disparity map from the coded image and center view 3) warping center view using the disparity to generate light field. We propose three neural networks for these stages. Our disparity estimation network is trained in an unsupervised manner alleviating the need for ground truth disparity. Our results demonstrate better recovery of parallax from the coded image and sharper reconstruction than dictionary learning approaches. All our results and code would be available at our project page \cite{clf2017}. 
   \vspace{-0.5cm}
\end{abstract}
\section{Introduction}
Imaging technologies have been striving to capture the rich three-dimensional scene around us as it is. But, since ages conventional cameras have only been able to project the scene to a two-dimensional photograph. Light field imaging is a step closer in this direction. Light field captures different perspective shifts of the same scene. It enables post capture facilities like refocusing and view-point changes. This is realized by jointly modifying the camera optics and computationally processing the  recorded data. The advent of light field imaging technology into affordable commercial cameras like Lytro \cite{lytro2017} has renewed research interests resulting its applications in areas like cinematography, 3D imaging, AR and VR etc.

Many methods have been proposed for capturing light field data \cite{levoy1996light, wilburn2005high, adelson1992single, ng2005light, georgiev2006spatio}. Of these, micro-lens array based acquisition by Ng et al. \cite{ng2005light} has successfully been adopted into lytro camera \cite{lytro2017}. Most of these methods suffer from a common problem. Due to limited sensor resolution trade-off arises between the spatial and angular resolution. These cameras have to sacrifice the spatial resolution to gain angular information. This hurts the megapixel-hungry trend with the consumer cameras. 

Recently, high resolution light field imaging has received a lot of attention from the research community. Early attempts used computer vision techniques to enhance spatio-angular resolution by exploiting the redundancy in 4D light field \cite{georgiev2006spatio}. Compressive light field imaging works backed by compressive sensing (CS) theory attempt to reconstruct full sensor resolution light field either from a set of a coded images \cite{babacan2012compressive} or a single image \cite{marwah2013compressive}. Since the reconstruction is ill-posed, data priors are essential. With the success of deep learning techniques for low-level image processing, recently, some works have proposed to use it for high resolution light field imaging \cite{kalantari2016learning, yoon2015learning}. 

In this work, we propose to use a deep neural network for reconstructing high resolution light field $\nu \times \nu \times h \times w$ from a single coded image $h\times w$. Our pipeline involves 1) center view reconstruction from the coded image 2) disparity map estimation from the center view and coded image and 3) warping center view using the disparity map to reconstruct light field. We show our results using simulations on Lytro illum dataset by Kalantari et al. \cite{kalantari2016learning}. Our major contributions are as follows:
\begin{itemize}
    \item We propose a deep learning based method for full sensor resolution light field reconstruction from a single coded image. 
    \item Output of our second network is disparity which in general would require ground truth disparity for training. Following the recent unsupervised learning works \cite{godard2016unsupervised} we learn it in an unsupervised manner. 
    \item Our disparity map based view synthesis inherently respects the parallax, depth dependent disparity in light field image. In contrast, a direct regression to light field from a coded image is not guaranteed to do so.
\end{itemize}

\section{Related Work}

\textbf{Light Field Imaging:} Light field imaging is a century old, initial attempts used a pinhole array in front of sensor to capture it,  \cite{lippmann}. Later people used camera arrays and other hardware modifications  \cite{levoy1996light, georgiev2006spatio}. Being bulky and expensive they were not affordable for commercial use. Recently, lenset based light field imaging \cite{ng2005light} has been integrated into commercial light cameras; Lytro \cite{lytro2017} and Pelican Imaging \cite{pelican2017}. This has renewed their application in 3D imaging. However, these lenslet cameras suffer from a common problem of spatio-angular tradeoff. They have to sparsely sample in either domain. For example, Pelican cameras have a $2\times 2$ array.

\textbf{High-resolution Light Field Imaging:} Given the trade-off between spatial-angular resolution many methods have been proposed to address the problem. Broadly these can be categorized in two categories. First set of methods sample the light field sparsely in one of the domains and post process using computer vision techniques to enhance spatio-angular resolution. Mitra et al. \cite{mitra2012light} proposed GMM light field prior for light field super-resolution, Bishop et al. \cite{bishop2009light} propose a variational bayes framework for the same. 

Second set of methods propose computational imaging based solutions. Levin et al. \cite{levin2010linear} sythesize 4D light field from 3D focal stack. Backed by compressive sensing, Marwah et al. \cite{marwah2013compressive} recover full sensor resolution light field from a single coded image. Babacan et al. \cite{babacan2012compressive} propose to reconstruct light field to reconstruct from a set of compressive measurements. 

\textbf{Compressive Light Field Imaging:} Marwah et al. place a coded mask at a distance from sensor multiplexing the angular information to a coded image on the sensor. They use light field dictionary as a prior for reconstruction from the coded image. Instead, here we propose to use deep neural nets for reconstruction which are shown to outperform dictionary learning in many low-level image processing tasks (see figure \ref{dl_cnn_cmp}). 

\textbf{Deep Learning:} With the successful application of deep learning for low-level image processing. Learning based solutions have been proposed for first category of methods. Yoon et al. \cite{yoon2015learning} do light field super-resolution by a factor of two using CNNs. Recently, Kalantari et al. \cite{kalantari2016learning} used learning based methods for view synthesis. Particularly, they were able to show convincing results on challenging real scenes by generating full $8\times 8$ light field from only four corner views.  In this work, we propose a deep learning solution for second category of methods. A learning based approach to reconstruct light field from coded image.

\textbf{Unsupervised Disparity Estimation:} Recently, unsupervised methods have been proposed for learning depth maps from stereo image pairs without access to ground-truth depth maps \cite{flynn2016deepstereo, godard2016unsupervised}. Godard et al. \cite{godard2016unsupervised} use photometric losses between the warped left view and right view achieving state of the art results on KITTI dataset. Unlike these methods, we don't have access to the stereo pair; only the coded image is used to estimate the depth.

\begin{figure*}[t]
    \centering
    \includegraphics[trim={0 0 0 0}, width=6.8in]{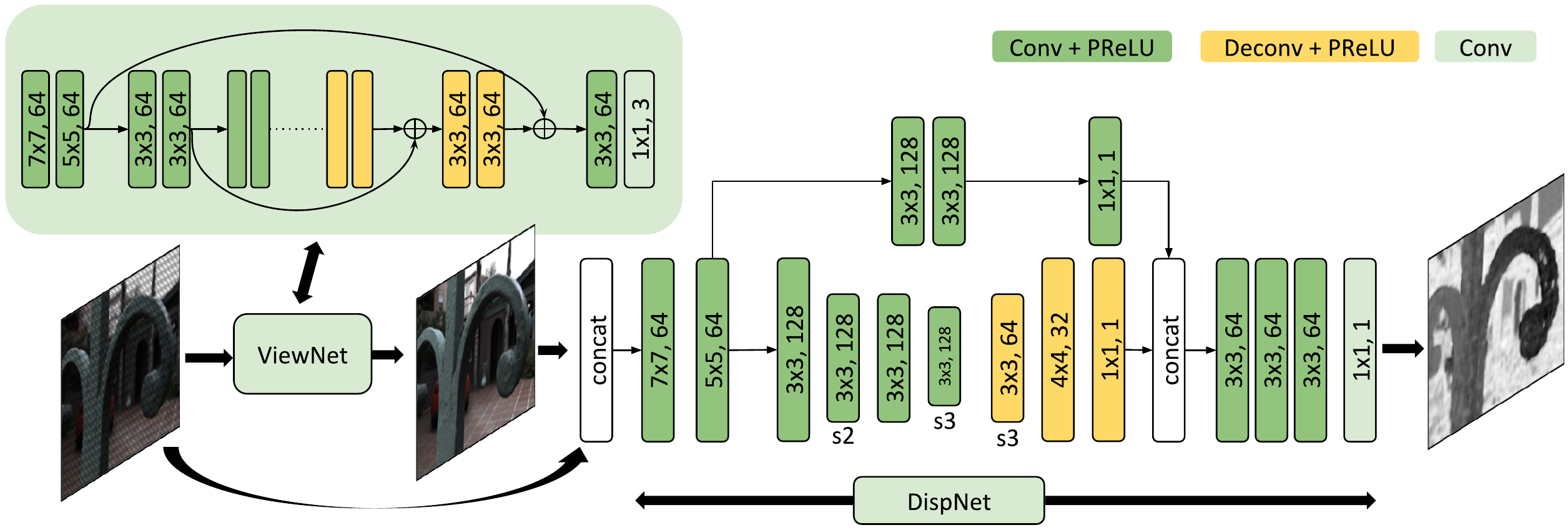}
    \caption{\textit{ViewNet} and \textit{DispNet}; figure shows the convolutional architecture details of both the networks. ViewNet is a convolutional deconvolutional architecture with three symmteric skip connections. It is shown compactly and can be easily inferred from Mao et al. \cite{mao2016image}. DispNet is an encoder decoder (bottom) with a skip connection (top). The `sk' specified below the block implies stride $k$ convolution. Each block shows the filter size ($c$) and number of output channels ($h$) as $c\times c, h$. }
    \label{dispnet}
\end{figure*}

\section{Proposed Method}
\subsection{Coded Image Simulation}
To simulate coded image we follow the same approach followed in \cite{marwah2013compressive}. Basically, this assumes a mask at a distance from sensor which optically modulates the incoming light field $L(x,\nu)$, projecting the coded image $I_c(x)$, on to the sensor. Mathematically this can be described as, 
\begin{equation}
    \label{clf}
    I_c(x) = \int_{\nu} f(x,\nu)L(x,\nu)  d\nu,
\end{equation}
where, $x$ is the spatial dimension on the sensor plane and $\nu$ is the angular dimension on the aperture plane. 
Given this coded image $I_c$, our goal is to recover the whole light field $L$, which is of the same resolution as the coded image. Thus, tackling the resolution trade-off. 

\subsection{DNNs for Light Field Reconstruction}
A trivial way of doing this is to train a CNN to directly regress for novel views. This might be inefficient for two reasons. First, light field exhibits depth dependent disparity (epi-polar geometry) and direct regression to novel views is not guaranteed to respect this. Secondly, as is noted by recent works on view synthesis \cite{flynn2016deepstereo, kalantari2016learning} training a neural network directly to warp/translate image is inefficient. Hence, we break down the light field reconstruction into two steps of scene reconstruction (center view) and scene geometry (disparity) estimation, then render full light field. Figure \ref{pipeline} shows the proposed pipeline for compressive light field reconstruction. 


In the following sections, we discuss the individual network architecture details. 

\subsubsection{Center View Reconstruction: ViewNet}
We need to reconstruct one of the views to synthesize the novel views. Intuitively, center view ($L_0$) of the light field is ideal for this. We use a series of convolutional layers for reconstructing center view, $L_0$ from the coded image $I_c$. The regions further away from the focal plane are proportionately degraded in $I_c$ as they see much parallax. To better reconstruct such details we incorporate symmetric skip connections proposed by Mao et al. \cite{mao2016image} (see figure \ref{dispnet}). 

\subsubsection{Learning Disparity Estimation: DispNet}
Once we have the center view reconstructed from ViewNet, we use both the coded image, $I_c$ and center view, $L_0$ to get the disparity map, $D$ corresponding to the center view. Now, this disparity map can be used to generate novel view at $q$, $\hat{L}_q$ from the center view $L_0$, under the Lambertian Scene assumption. This is mathematically given as,
\begin{equation}
    L_q(x + qD(x)) = L_0(x),
    \label{warping}
\end{equation}
where $x$ is the pixel location. This is same as the forward warping in case of stereo pair.  

Here, we propose a deep neural network, \textit{DispNet}, for disparity map estimation. We stack the coded image and reconstructed center view as input to DispNet. Center view is used to ensure that disparity maps aligns with it.

\textbf{DispNet} shown in figure \ref{dispnet} is used for disparity estimation. It consists of encoder-decoder architecture with a skip-connection. This network is designed along the lines of recent unsupervised disparity estimation networks \cite{godard2016unsupervised}. Encoder performs strided($>1$) convolutions encoding the disparity information. Also, this provides a way for increasing the receptive fields. Decoder performs strided deconvolutions bringing back the feature map to the input resolution. Skip connections ensure that the low level details are preserved in the disparity maps. The feature maps from decoder and skip connection are concatenated, followed by four convolutional layers to output final disparity map, $D$. This goes as input to the warping block. 

\subsubsection{Warping and Interpolation: WarpNet}
The disparity map, $D$ is used to synthesize the novel views by forward-warping shown in eq. \eqref{warping}. However, the presence of depth edges always introduces some occluded regions which can't be recovered by forward-warping. To tackle this, we use four more convolutional layers on top of the warped view for interpolation job. We refer to this network as \textit{WarpNet} and it outputs the final novel view. 
\begin{figure*}[!t]
\centering
    \hspace{.087cm}
    \begin{minipage}[t]{.3\linewidth}
        \includegraphics[width=1\linewidth]{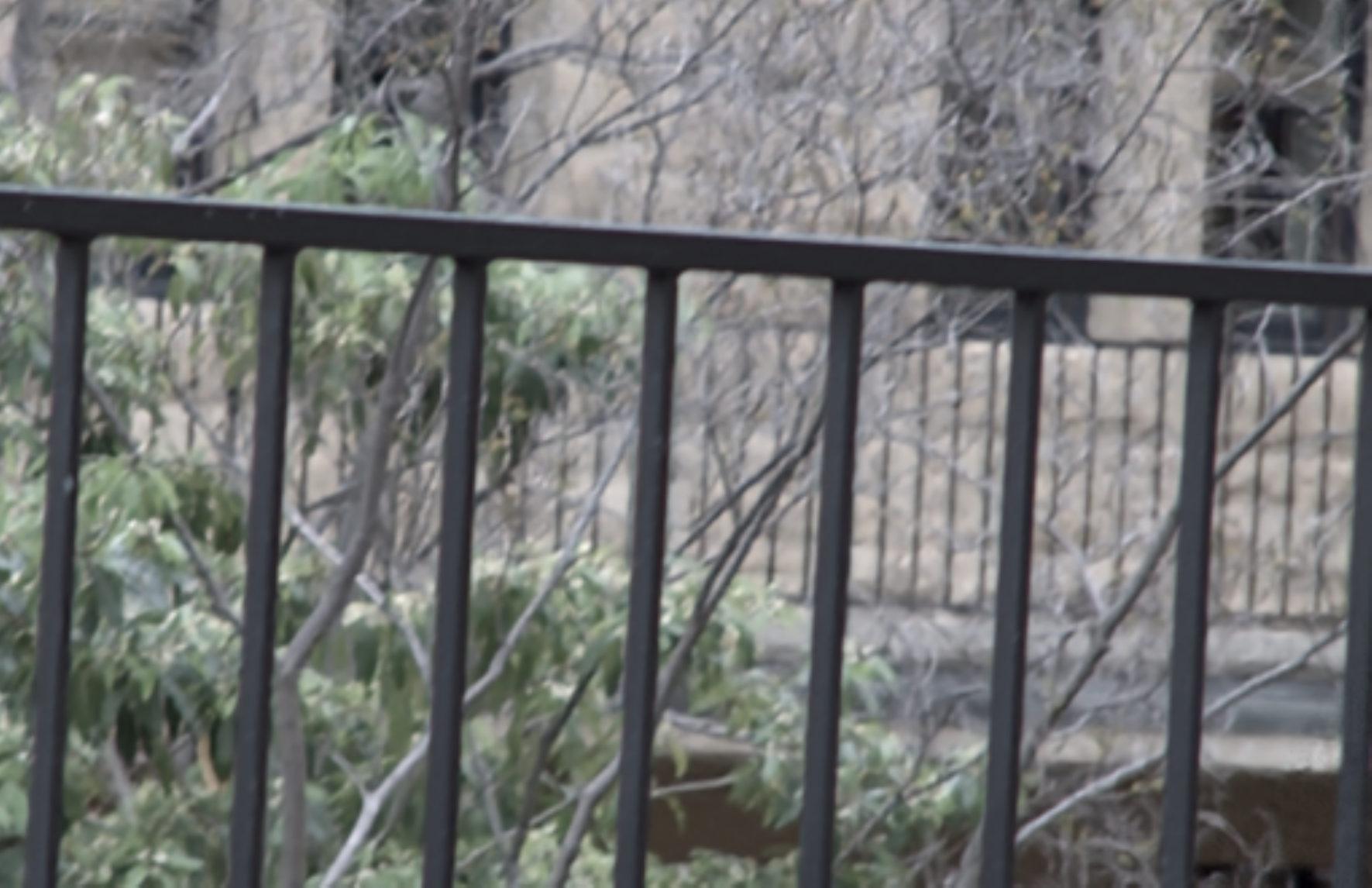}\\
        \includegraphics[width=1\linewidth]{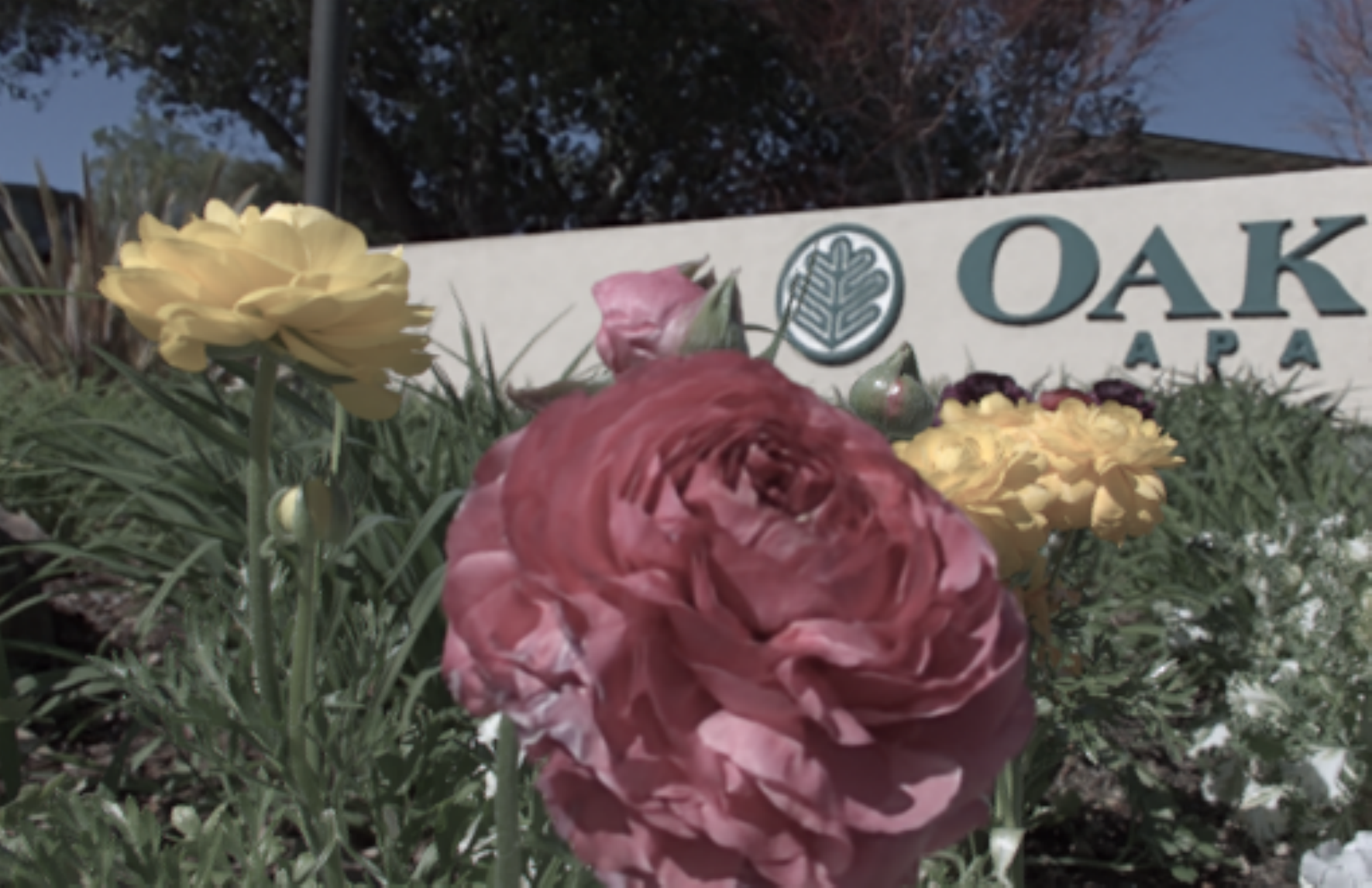}\\
        \centerline{\small Reconstructed center view}\\
    \end{minipage}\hspace{.15cm}
    \begin{minipage}[t]{.3\linewidth}
        \includegraphics[width=1\linewidth]{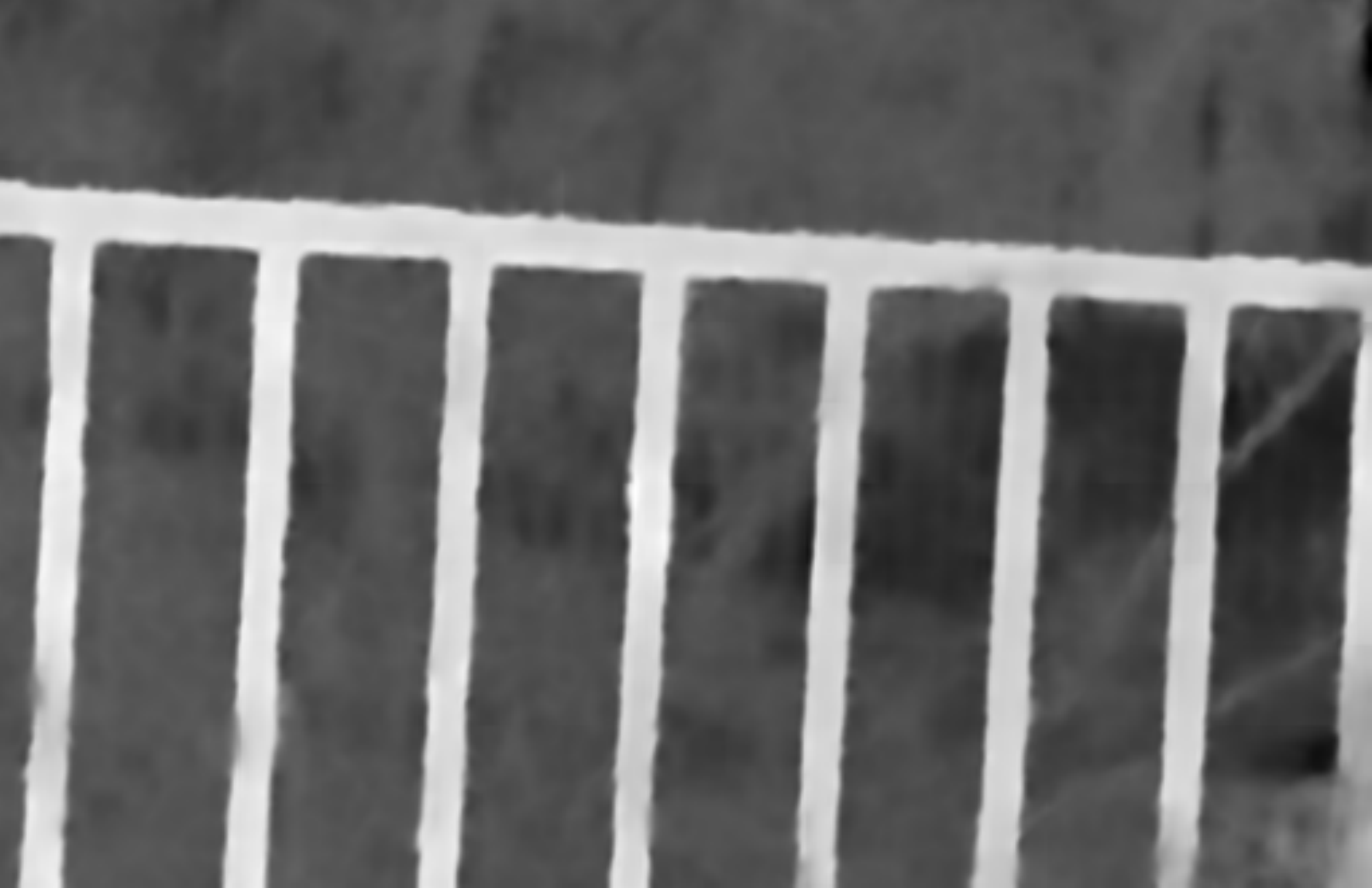}\\
        \includegraphics[width=1\linewidth]{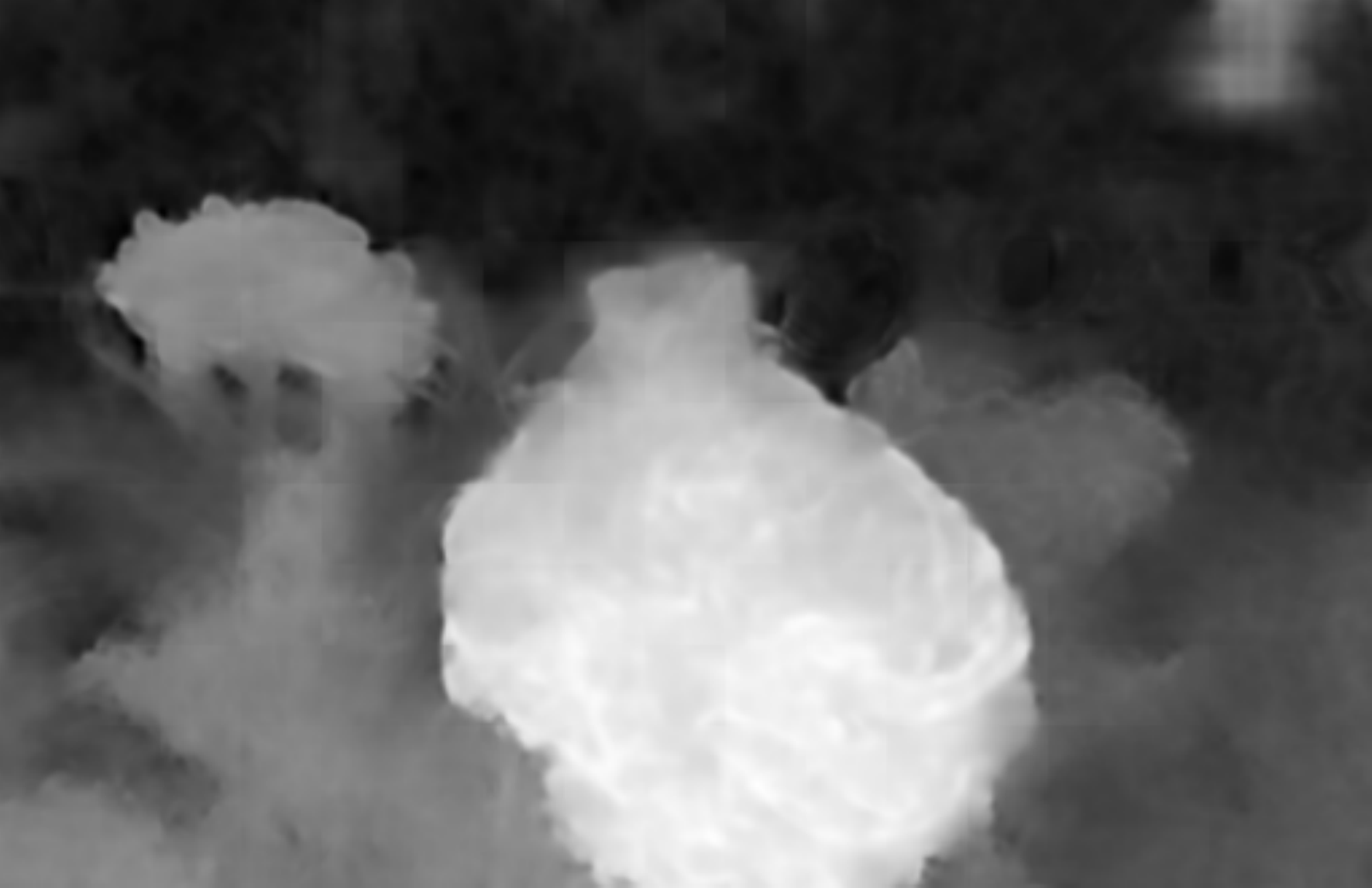}\\
        {\small Our disparity from the coded image and the reconstructed center view by DispNet} 
    \end{minipage}\hspace{.4cm}  
    \begin{minipage}[t]{.3\linewidth}
        \includegraphics[width=1\linewidth]{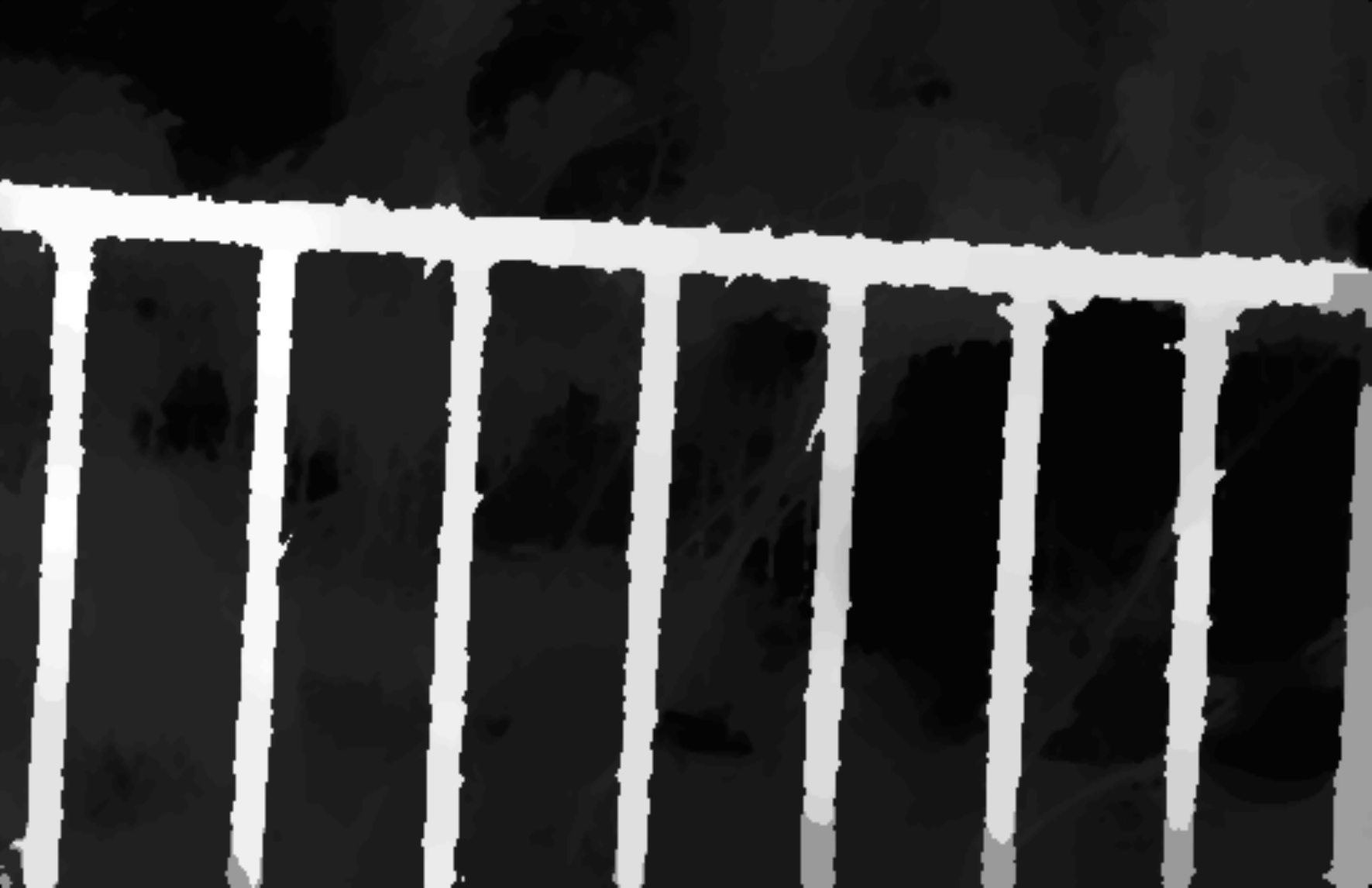}\\
        \includegraphics[width=1\linewidth]{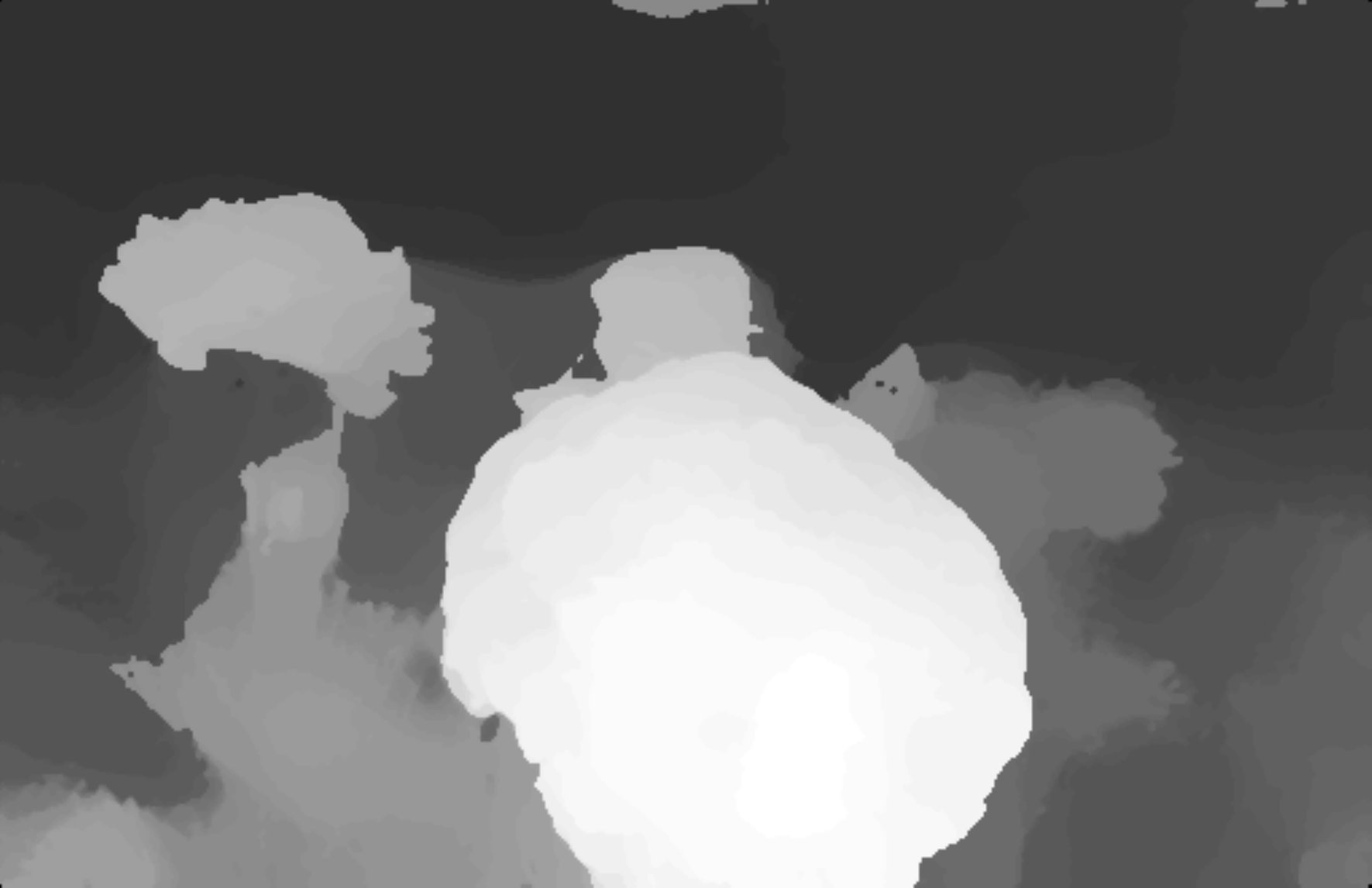}\\
        {\small Jeon et al. \cite{jeon2015accurate} disparity estimation using full light field}
    \end{minipage} 
    \caption{Disparity map comparison: Using only photo-consistency loss our network is able to learn disparity maps as good as one estimated from ground truth light field. Note that our estimation is without any explicit regularization unlike \cite{jeon2015accurate}. Please see our project page \cite{clf2017} for comparisons on a variety of scenes.}
    \label{disparity_cmp}
\end{figure*}

\begin{figure}[t]
    \centering
    \begin{minipage}{.1\textwidth}
        \vspace{0.053cm}
        \includegraphics[trim={0.3cm 0.3cm 0.3cm 0.3cm},clip,width=0.95\textwidth]{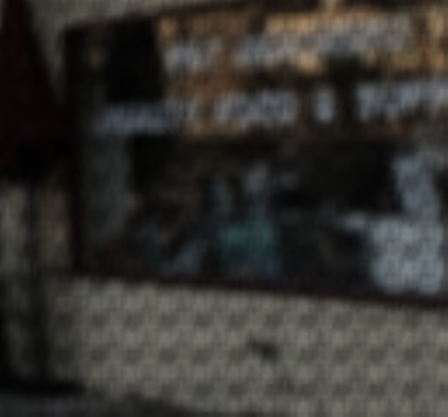}
    \end{minipage}        
    \begin{minipage}{.1\textwidth}        
        \includegraphics[trim={0.3cm 0.3cm 0.3cm 0.3cm},clip,width=0.95\textwidth]{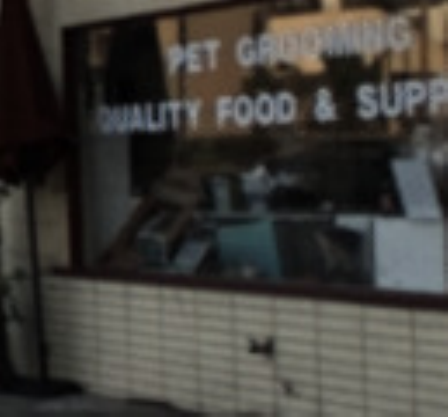}
    \end{minipage}  
    \begin{minipage}{.1\textwidth}
        \includegraphics[trim={0.3cm 0.3cm 0.3cm 0.3cm},clip,width=0.95\textwidth]{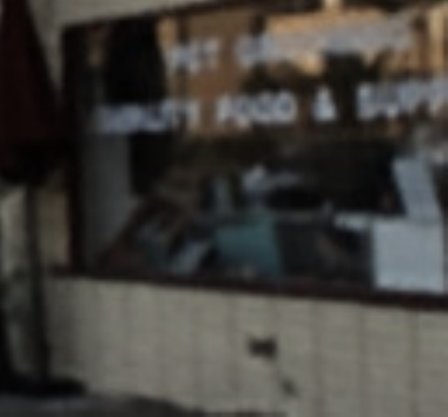}
    \end{minipage}                    
    \begin{minipage}{.1\textwidth}
        \vspace{0.053cm}
        \includegraphics[trim={0.3cm 0.3cm 0.3cm 0.3cm},clip,width=0.95\textwidth]{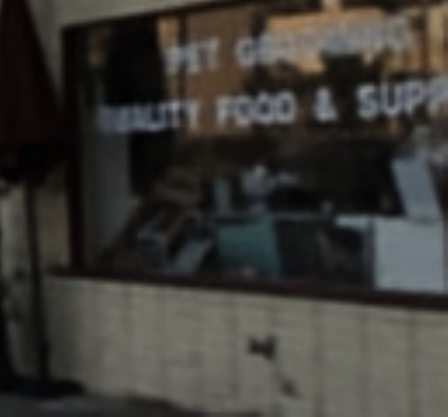} 
    \end{minipage}         
    \caption{Center view reconstruction with \textit{ViewNet}: Notice recovery of sharp details like text over the glass and pattern edges below it with Conv-Deconv network. Conv-Deconv network employs symmetric skip connections proposed by Mao et al.}
    \label{cv_reconst}
\end{figure}

\section{Training}
Finally, the generated view is compared with ground truth and loss is back-propagated through the warping and disparity estimator network. The loss is given as, 
\begin{equation}
    \label{loss}
    E = \sum_k (\hat{L}_{q,k} - L_{q,k})^2, 
\end{equation}
where sum is over RGB color channels. The generated view at $q$, $\hat{L}_q$ is a function of parameters of both disparity estimator $(W_d)$ and warping network $(W_w)$.   

\begin{equation}
    \label{warp_grad}
    \displaystyle \frac{\partial E}{\partial W_d} = \sum_{k} \left(\frac{\partial E}{\partial \hat{L}_{q,k}} \frac{\partial \hat{L}_{q,k}}{\partial D} \right) \frac{\partial D}{\partial W_d}
\end{equation}

For training using gradient descent we need the gradients of the loss \eqref{loss} with respect to the parameters ${W_d, W_w}$. The gradients ${\partial E}/{\partial W_d}$ involves estimating the second term of equation \eqref{warp_grad} which is gradient of warping function with dispartiy map, $D$. Although bicubic interpolation is differentiable numerical estimation of gradients gives satisfactory results. Other parameter gradients can be obtained using standard back-propagation techniques. 


\subsection{Implementation Details}
\label{implementation}
For training, we use the 100 Lytro illum images provided by Kalantari et al. \cite{kalantari2016learning}. We split this data randomly into 85 training images and 15 validation images. We extract 2 million light field patches  ($7\times7\times120\times120$) from the training data. We used random Gaussian code $15\times 15$, clipped between 0 and 1 to simulate coded image ($120 \times 120$). We use caffe \cite{jia2014caffe} to train with Adam optimizer \cite{kingma2014adam}. We set initial learning rate $1e-4$ and reduce it by a factor of 0.8 every 5 epochs. Except with the intial $7\times 7,~ 5\times 5$ convolutional layers we do padded convolutions to preserve the input-ouput size. To accommodate for the loss in spatial resolution due to initial unpadded convolutions we appropriately crop the label ($110\times 110$) to evaluate the loss. 

Initially, we pretrain the individual blocks separately and then combine them for an end-end training. ViewNet is individually trained for center view reconstruction. DispNet and WarpNet are trained together with ground truth center view initially. Pretraining is done for 30 epochs. In end-end training, the ViewNet output is used as input to the DispNet and warping function. This joint optimization is carried on for 5 epochs.

\begin{figure*}[t!]
\centering
    \begin{minipage}{.95\textwidth}
       \includegraphics[width=1\textwidth]{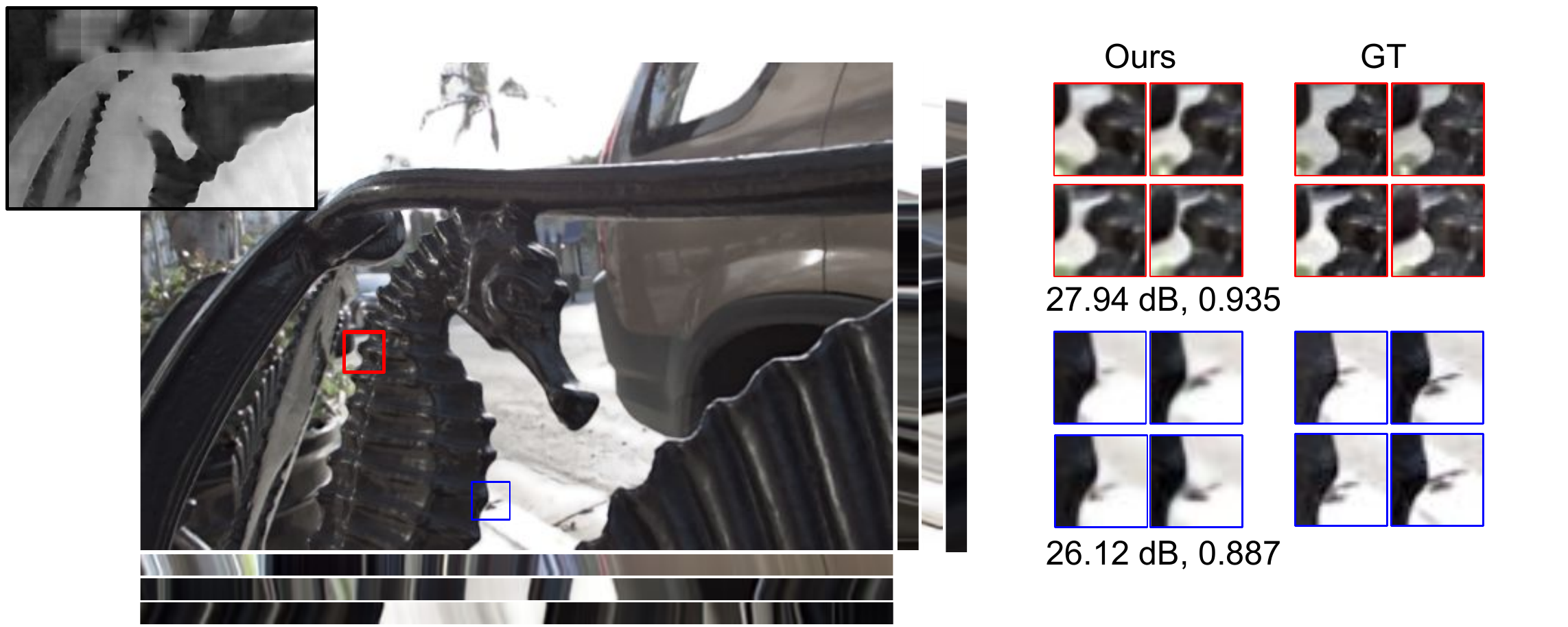}
       \centerline{Seahorse  scene - 33.16 dB, 0.953}
    \end{minipage}\\
    \vspace{0.2cm}
    \begin{minipage}{.45\textwidth}
       \includegraphics[width=1\textwidth]{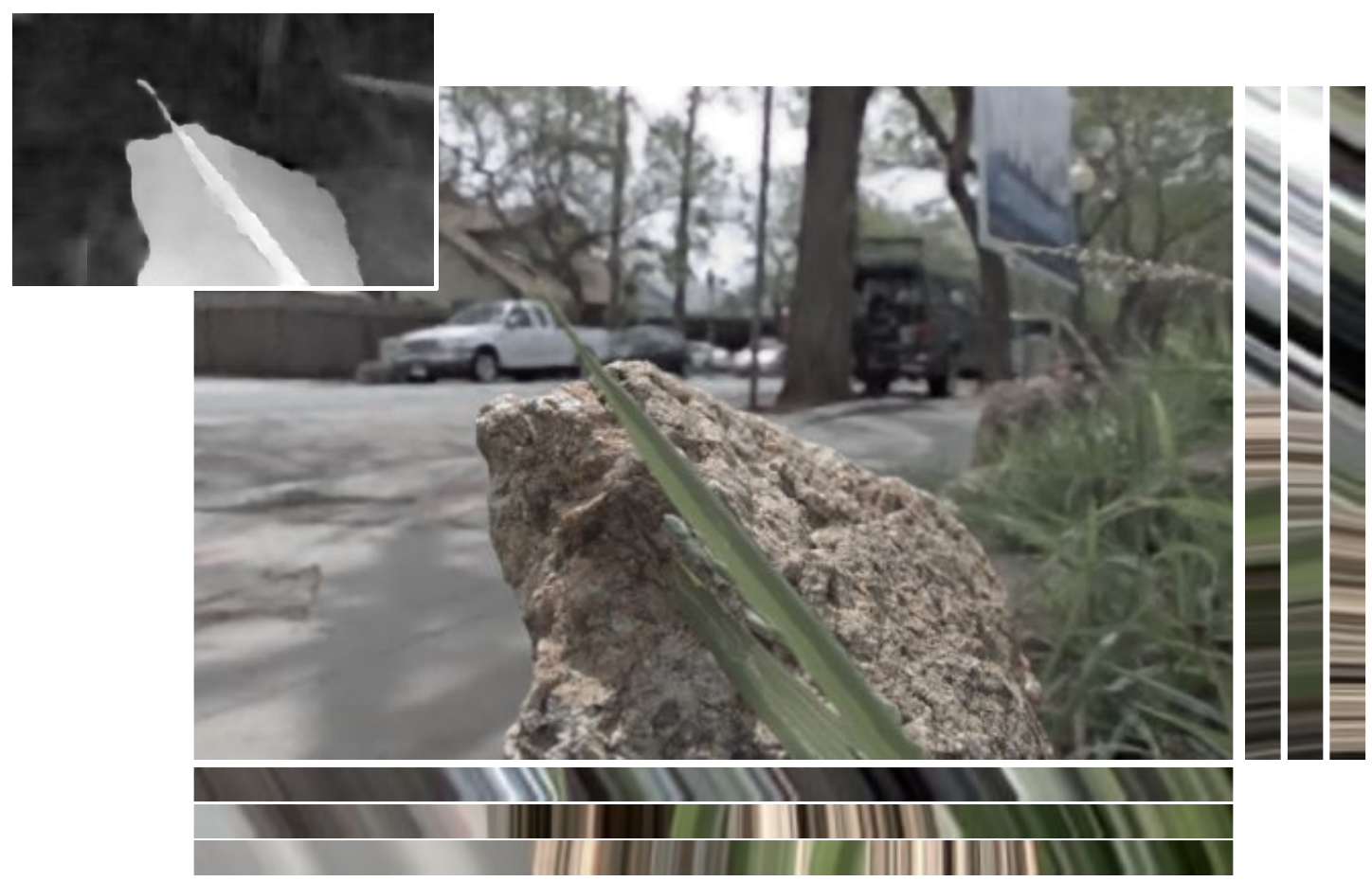}        \centerline{Rock scene - 33.14 dB, 0.941}
    \end{minipage}    
    \begin{minipage}{.45\textwidth}
       \includegraphics[width=0.97\textwidth]{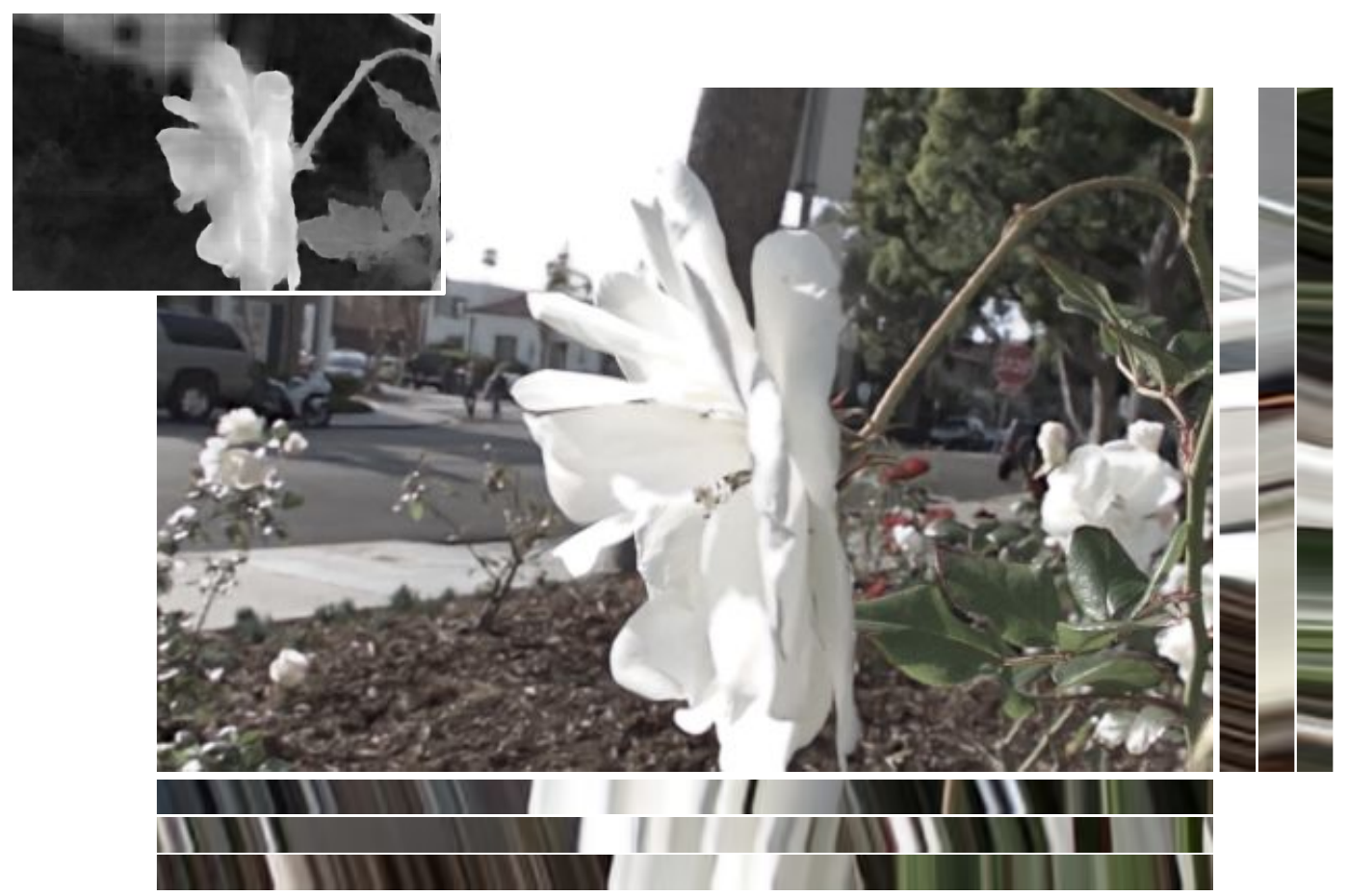}
       \centerline{FLowers2 scene - 31.67 dB, 0.942}
    \end{minipage}  
      \caption{Full sensor resolution light field reconstruction from coded image for different test scenes from Kalantari et al. \cite{kalantari2016learning}: figures show the reconstructed center view overlaid with the estimated disparity maps (brigter is closest). The numbers reported are PSNR in dB and SSIM. From the slope of EPI images we can see that our approach recovers parallax well also implied from the sharpness of the disparity maps at occlusion edges. This can also be noticed from the insets (blue and red) showing the corner views of 7x7 light field of Seahorse scene. We can see that ViewNet recovers fine details like the shadow pattern in the blue inset. Please check our project page \cite{clf2017} for animations of generated lightfield clearly showing the parallax.}
    \label{result_lf}
\end{figure*}

\section{Results}
\textbf{Center View Reconstruction} - As shown in figure \ref{cv_reconst}, adding skip connections in viewnet helps in better recovery of finer details over a series of convolutional layers. Notice the details like the text on glass and objects behind the glass.  

\textbf{Disparity Estimation} - Figure \ref{disparity_cmp} shows DispNet estimation when trained with coded image and ground truth center view. We compare our disparity map with one estimated from full light field by Jeon et al. \cite{jeon2015accurate}. We can see that our method estimates disparities as good as them with out any explicit regularization. Figure \ref{result_lf} shows the disparity maps estimated from the coded image using DispNet. Note that we don't use any ground truth disparity for training DispNet. We use the photoconsistency loss to learn the disparity map as has been recently employed for unsupervised disparity estimation \cite{godard2016unsupervised}. From figure \ref{result_lf}, we can see fine details in the disparity and its sharp around the occlusion edges, thus enabling better parallax recovery in light field. 

\textbf{Compressive Light Field Reconstruction} - Figure \ref{result_lf} shows the results of light field reconstruction at 7x7 angular resolution from the coded image using our pipeline. It shows the reconstructed center view along with estimated disparity overlaid on it. We can see that reconstructed EPI are sharp. The mentioned PSNR and SSIM values are the average across all the 49 reconstructed views. Note the parallax recovery also shown in insets for the Seahorse scene. 

\begin{figure*}[!t]
    \centering
    \includegraphics[width=0.93\textwidth]{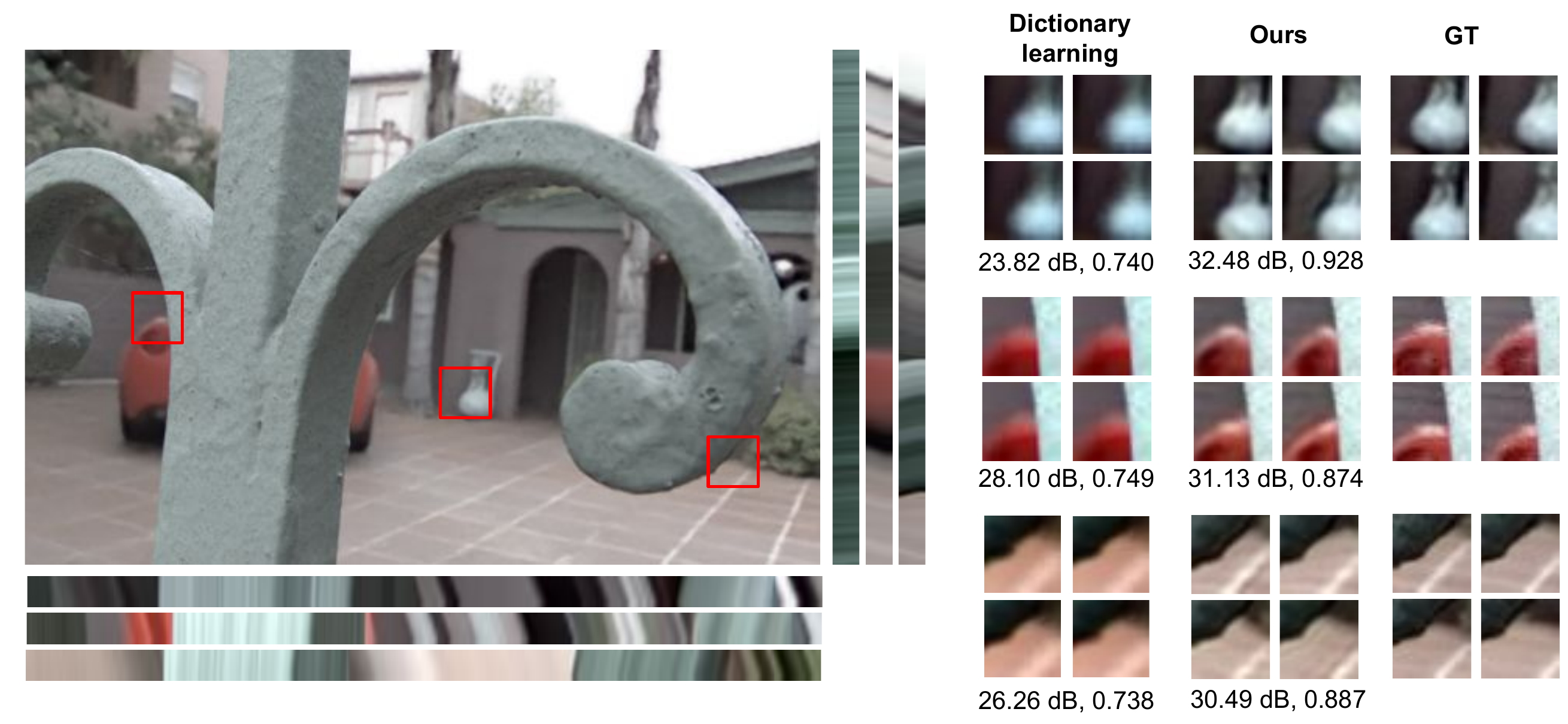}
    \caption{Comparison with dictionary learning: figure shows the center view reconstructed from our approach and EPIs from the light field (5x5). For the patches indicated in the figure, we compare our reconstructions with that dictionary learning based method of Marwah et al. \cite{marwah2013compressive}. As can be seen, our method recovers the parallax well and sharper reconstruction.}
    \label{dl_cnn_cmp}
\end{figure*}

\begin{figure*}[t]
    \centering
    \includegraphics[width=0.24\textwidth]{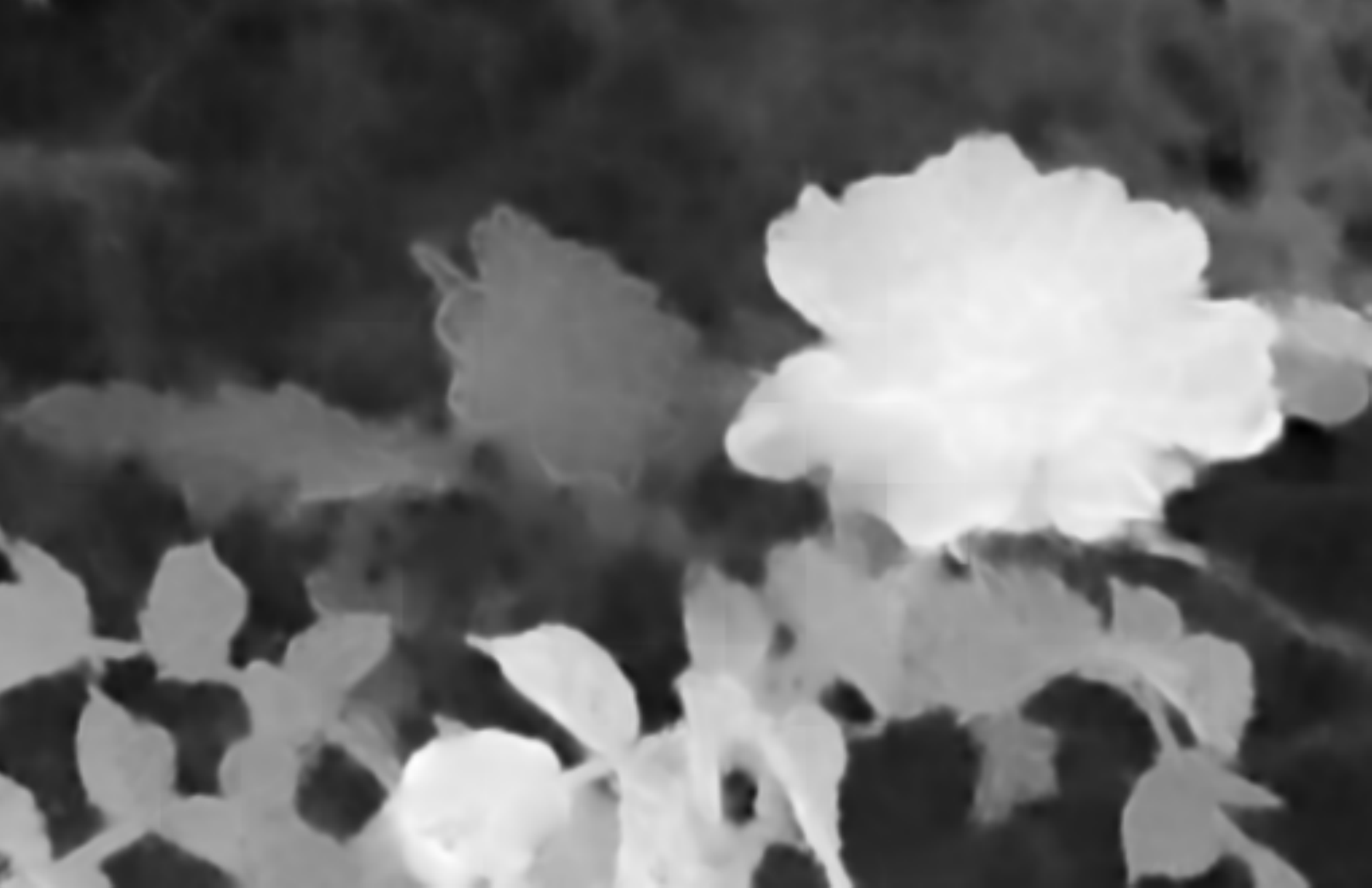}
    \includegraphics[width=0.24\textwidth]{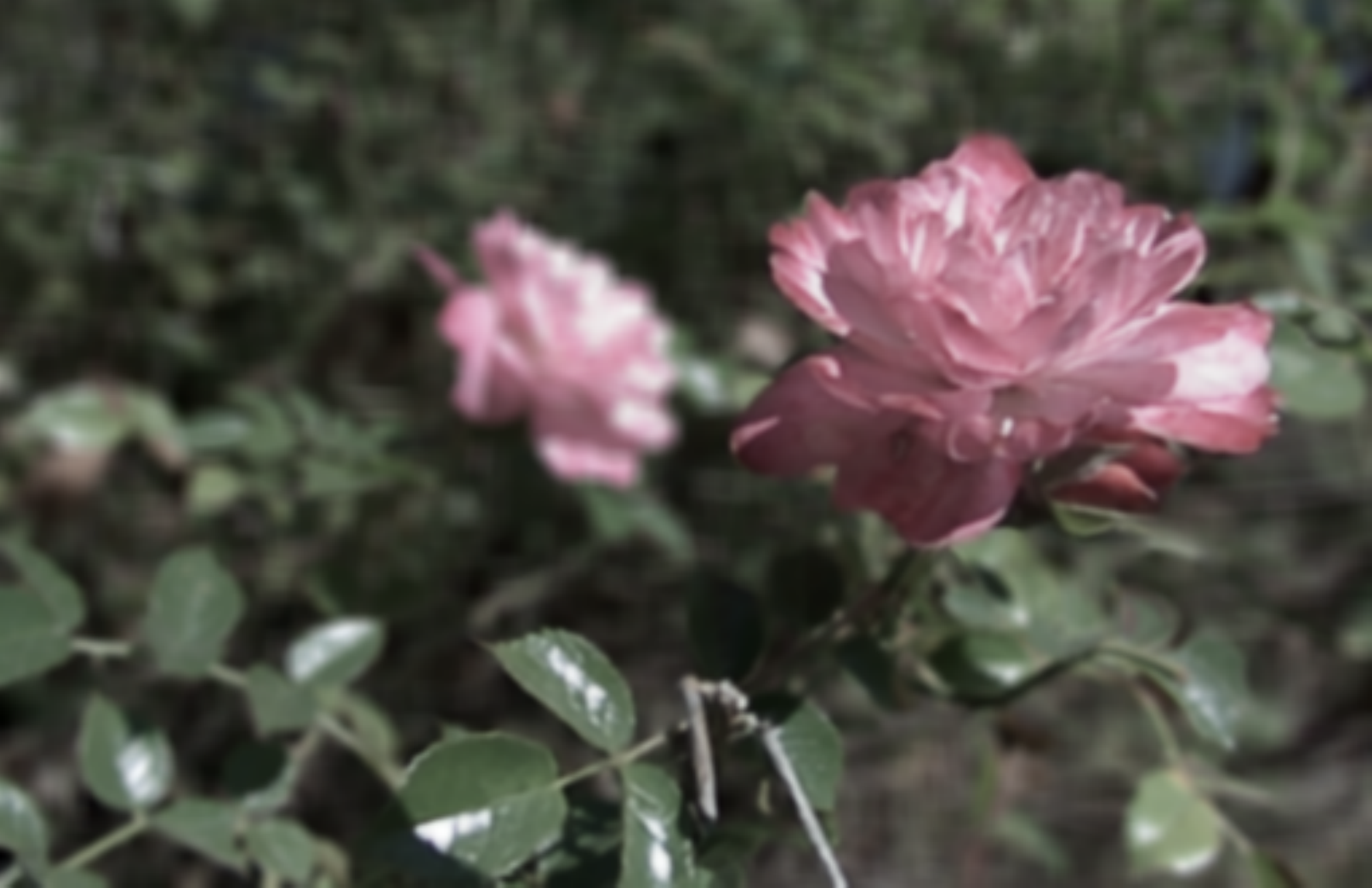}
    \includegraphics[width=0.24\textwidth]{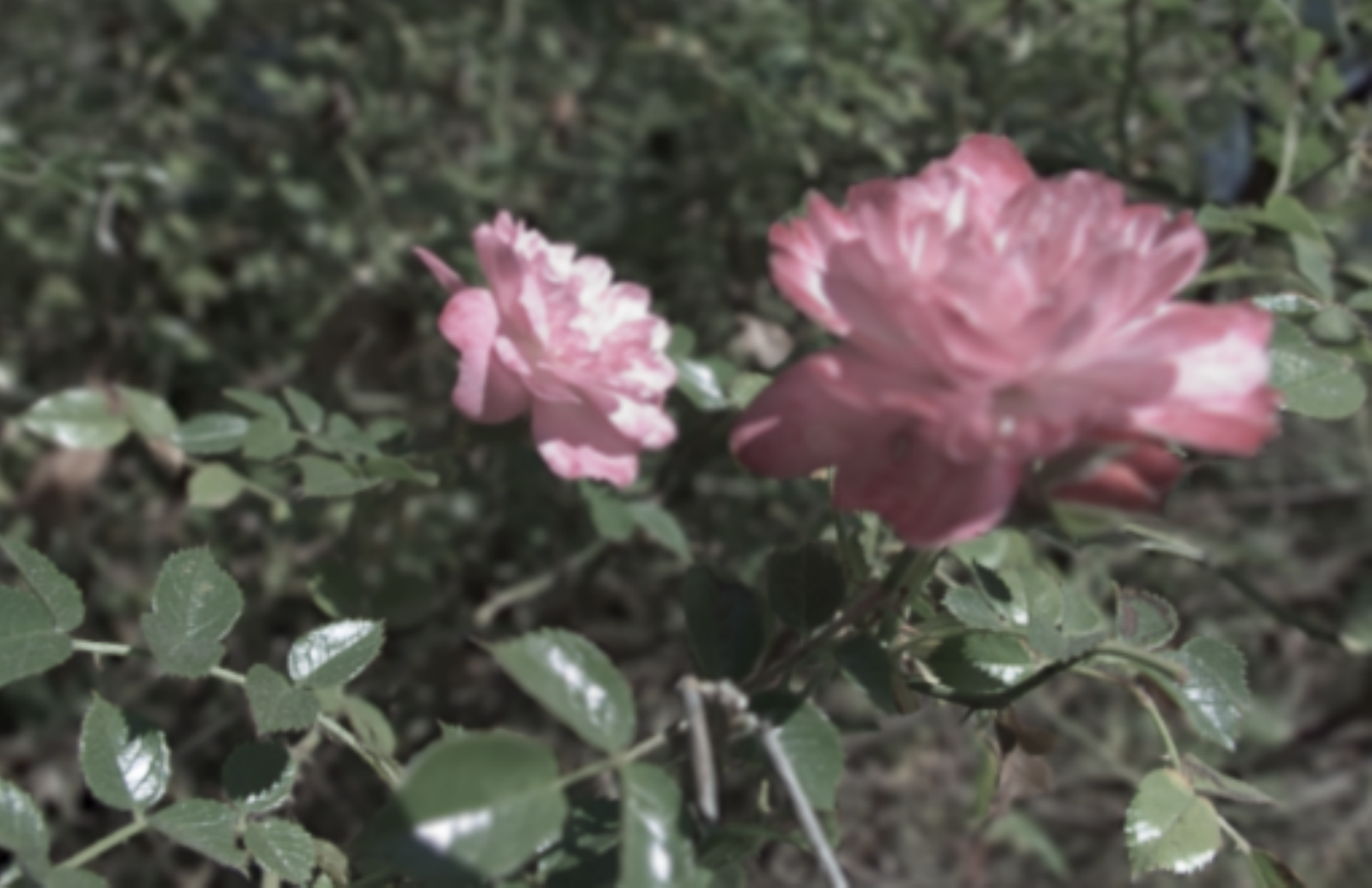}
    \includegraphics[width=0.24\textwidth]{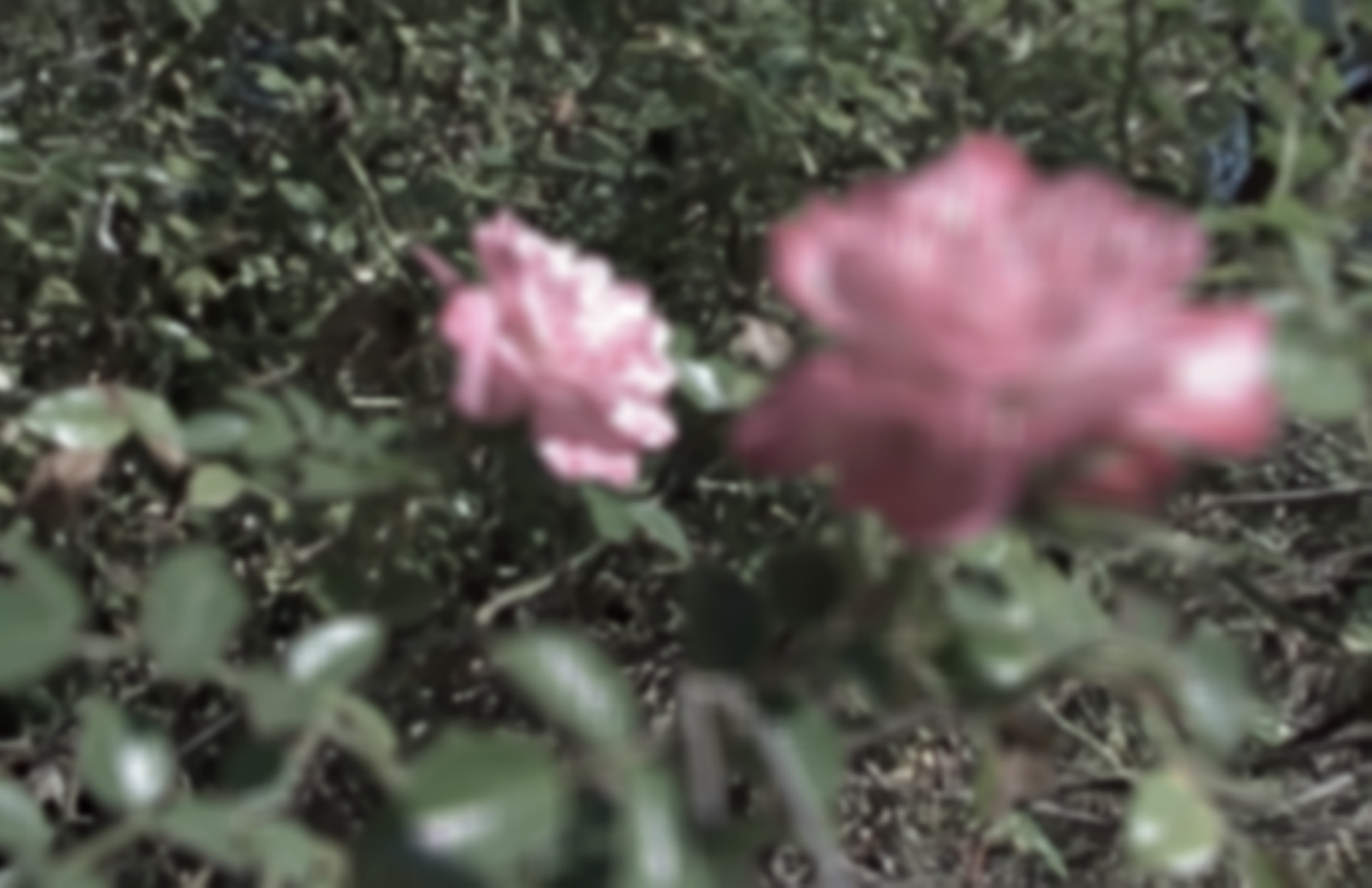}\\
    \vspace{0.2cm}
    \includegraphics[width=0.24\textwidth]{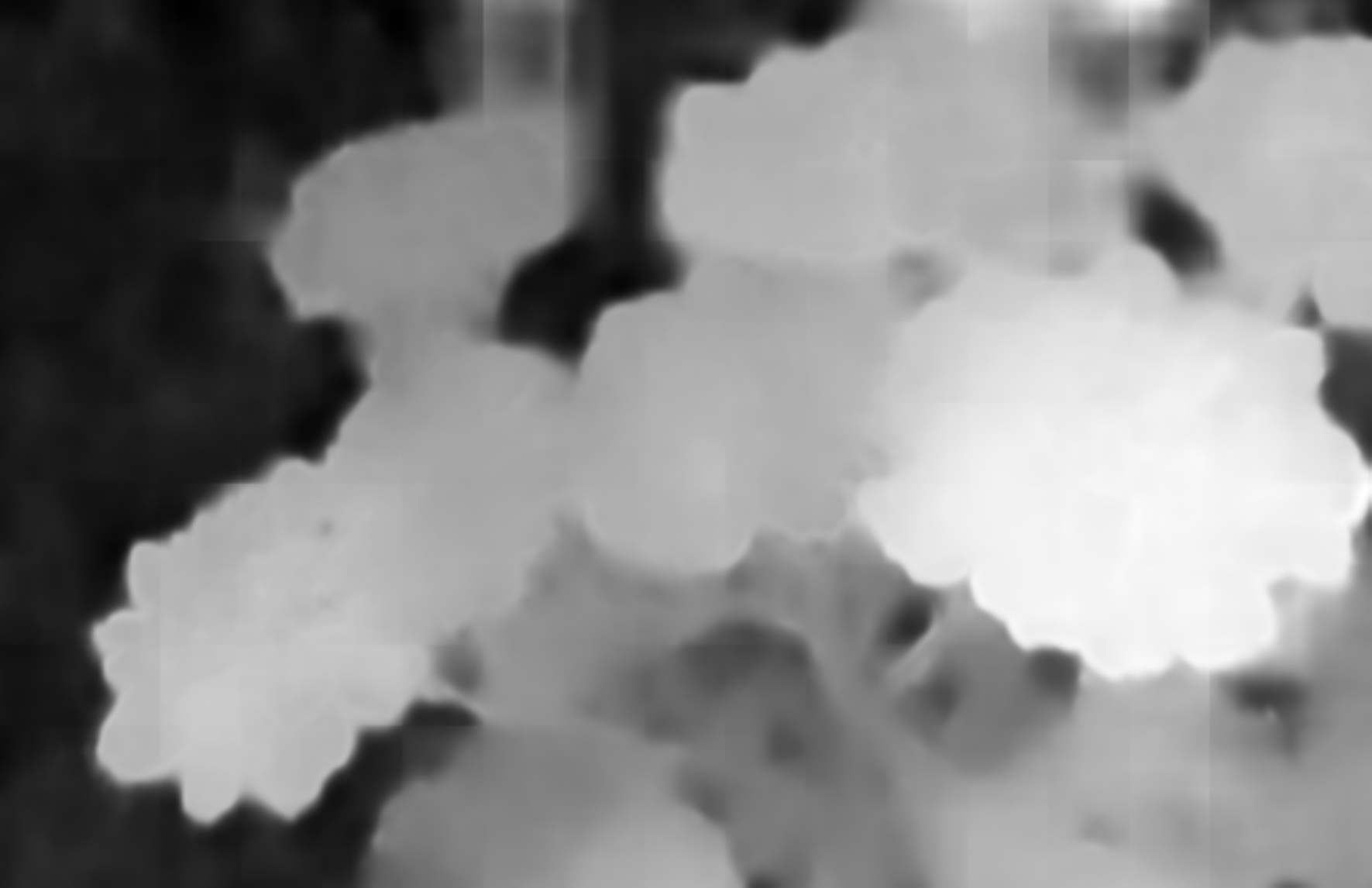}
    \includegraphics[width=0.24\textwidth]{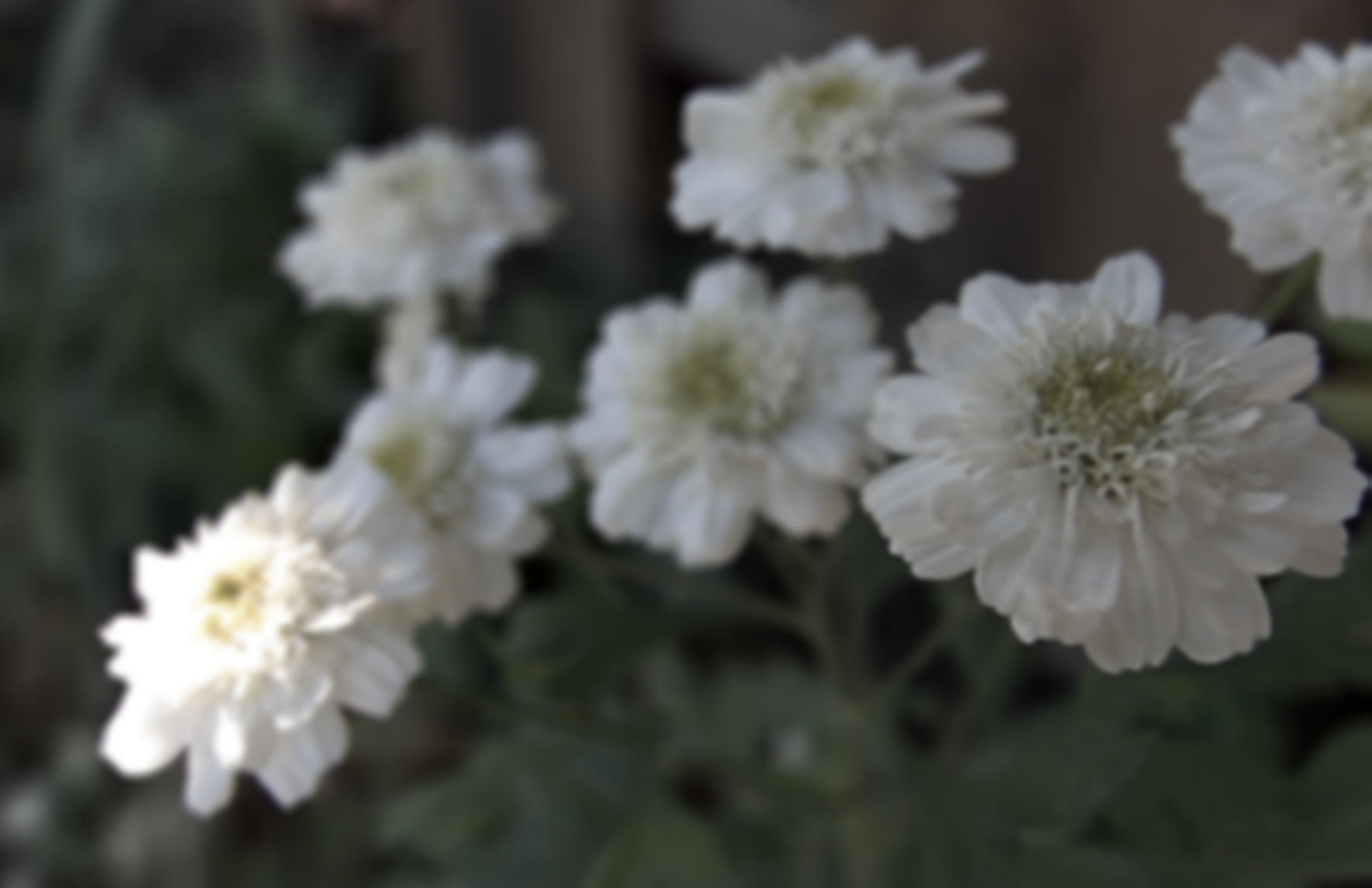}
    \includegraphics[width=0.24\textwidth]{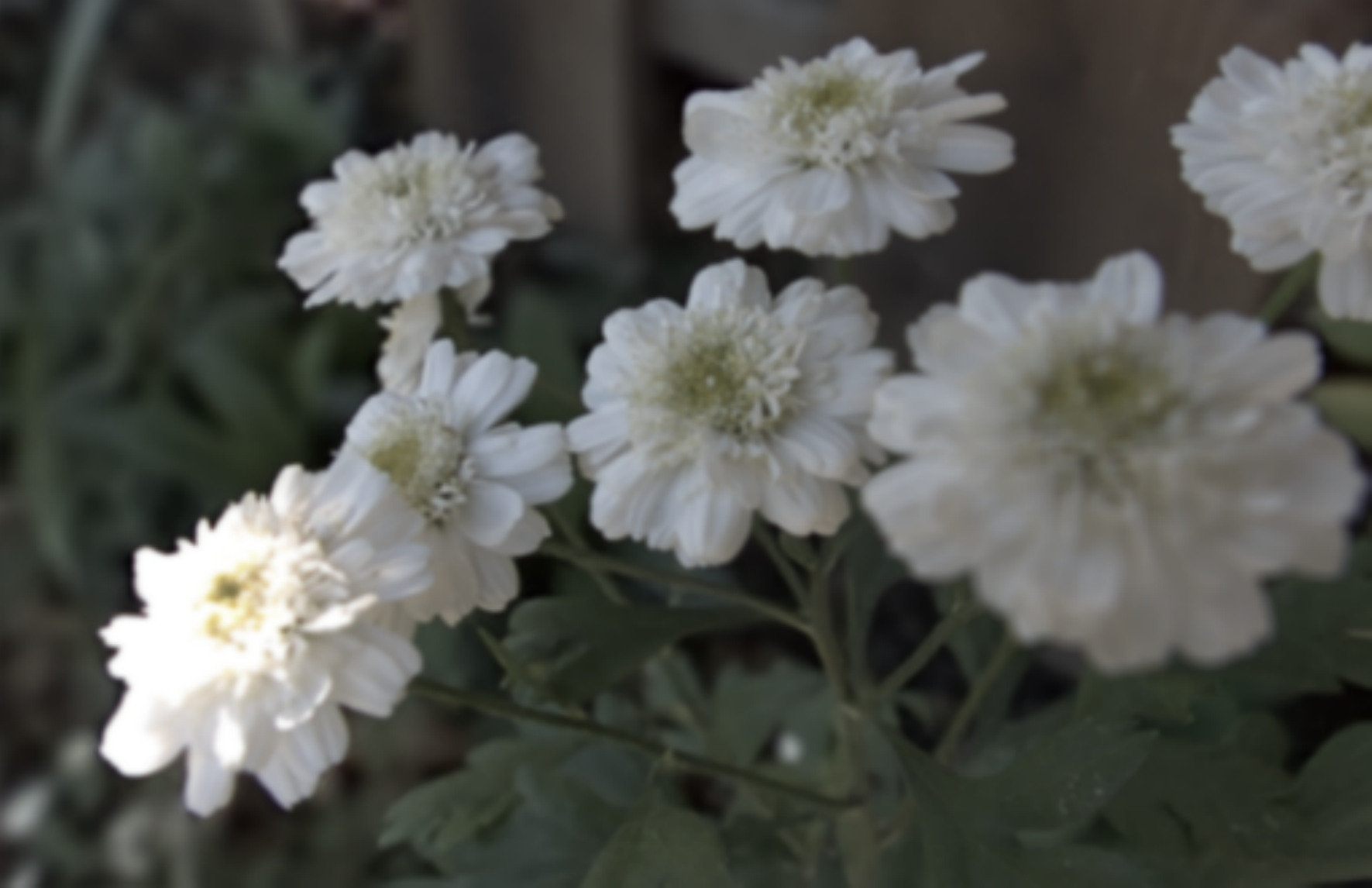}
    \includegraphics[width=0.24\textwidth]{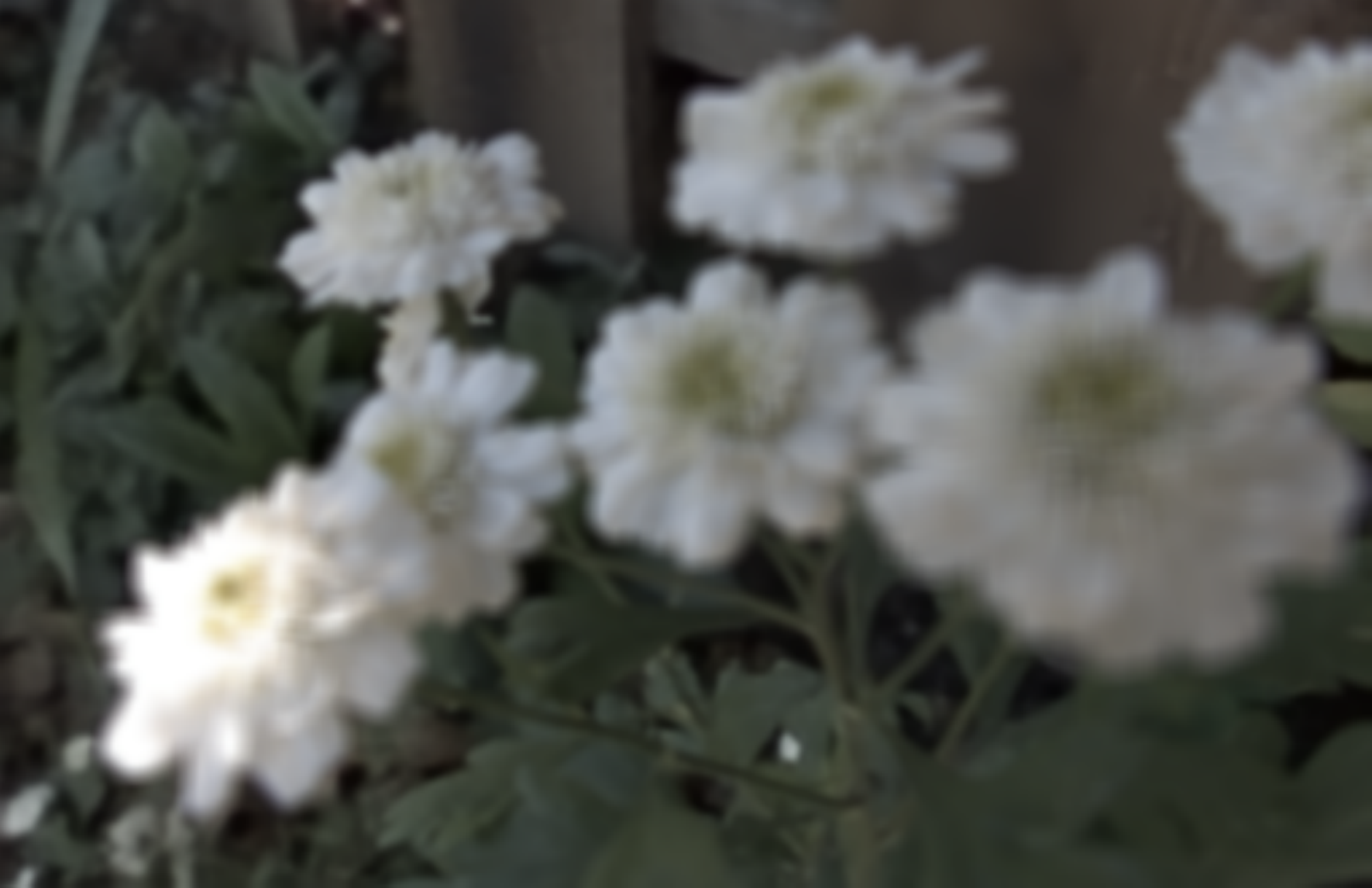}
    \caption{Refocusing application: The Flowers scene is refocused using the estimated disparity and generated light filed. The focus shifts from front to back from left to right. Note that refocusing is done by shifting and adding the light field images not by disparity based blurring.}
    \label{refocusing}
\end{figure*}

\textbf{Comparison with Dictionary Learning} - We compare our learning based light field reconstruction to that of dictionary learning based method of Marwah et al. \cite{marwah2013compressive}. To reduce computational burden here we compare our light field synthesis at $5 \times 5$ angular resolution instead of $7 \times 7$. We retrain our network for this case. For dictionary learning, we extract $5 \times 5 \times 11 \times 11$ light field patches from 20 illum training images. After coreset reduction we obtain 40,000 patches for training. We use SPAMS toolbox for learning dictionary \cite{mairal2014sparse} and perform reconstruction along the lines of Marwah et al. Figure \ref{dl_cnn_cmp} shows comparisons with our method. We can clearly see that our approach recovers parallax much better than dictionary learning. Also, reconstructions are sharper especially when the objects are farther from the focal plane like the pot in the background (top patch), the grid lines on the floor (bottom patch).

\textbf{Refocusing} - As an application of the reconstructed light field we show refocusing of the \textit{Flowers} scene using the generated disparity map (see figure \ref{refocusing}).

\section{Conclusions and Future work}
In this work we propose a learning based solution for sensor resolution preserving light field reconstruction from a single coded image. Our approach with explicit disparity based reconstruction is suitable for parallax recovery in the light field. We learn disparity estimation in an unsupervised manner. Our method performs better than dictionary based methods. In future, we can add explicit regularization \cite{godard2016unsupervised} to improve the disparity maps. The hardware realization of this work using the same setup of Marwah et al. Also, we can look at light field generation from a single RGB image.

{\small
\bibliographystyle{ieee}
\bibliography{clf_bib}
}

\end{document}